\documentclass[conference]{IEEEtran}
\IEEEoverridecommandlockouts
\usepackage{cite}
\usepackage{amsmath,amssymb,amsfonts}
\usepackage{algorithmic}
\usepackage{graphicx}
\usepackage{textcomp}
\usepackage{xcolor}

\usepackage{tabularx}
\usepackage{booktabs}
\usepackage{enumitem}
\usepackage{url} 
\DeclareRobustCommand*{\IEEEauthorrefmark}[1]{%
    \raisebox{0pt}[0pt][0pt]{\textsuperscript{\footnotesize\ensuremath{#1}}}}

\def\BibTeX{{\rm B\kern-.05em{\sc i\kern-.025em b}\kern-.08em
    T\kern-.1667em\lower.7ex\hbox{E}\kern-.125emX}}
\begin{document}

\title{


Molecular Graph Representation Learning Integrating Large Language Models with Domain-specific Small Models

}

\author{\IEEEauthorblockN{Tianyu Zhang\IEEEauthorrefmark{1},
Yuxiang Ren\IEEEauthorrefmark{2}$^\dagger$,
Chengbin Hou\IEEEauthorrefmark{3},  
Hairong Lv\IEEEauthorrefmark{1}$^\dagger$, and
Xuegong Zhang\IEEEauthorrefmark{1}}
\IEEEauthorblockA{\IEEEauthorrefmark{1}Department of Automation, Tsinghua University, Beijing, China}
\IEEEauthorblockA{\IEEEauthorrefmark{2}Huawei Technologies, China}
\IEEEauthorblockA{\IEEEauthorrefmark{3}School of Computer Science and Engineering, Fuyao University of Science and Technology, Fuzhou, China}
\thanks{$^\dagger$Corresponding authors: Yuxiang Ren; Hairong Lv.

E-mails: renyuxiang1@huawei.com; lvhairong@tsinghua.edu.cn.

Tianyu Zhang and Yuxiang Ren contribute equally to this paper.}
}

\maketitle

\begin{abstract}
Molecular property prediction is a crucial foundation for drug discovery. In recent years, pre-trained deep learning models have been widely applied to this task. Some approaches that incorporate prior biological domain knowledge into the pre-training framework have achieved impressive results. However, these methods heavily rely on biochemical experts, and retrieving and summarizing vast amounts of domain knowledge literature is both time-consuming and expensive. Large Language Models (LLMs) have demonstrated remarkable performance in understanding and efficiently providing general knowledge. Nevertheless, they occasionally exhibit hallucinations and lack precision in generating domain-specific knowledge. Conversely, Domain-specific Small Models (DSMs) possess rich domain knowledge and can accurately calculate molecular domain-related metrics. However, due to their limited model size and singular functionality, they lack the breadth of knowledge necessary for comprehensive representation learning. To leverage the advantages of both approaches in molecular property prediction, we propose a novel Molecular Graph representation learning framework that integrates Large language models and Domain-specific small models (MolGraph-LarDo). 
Technically, we design a two-stage prompt strategy where DSMs are introduced to calibrate the knowledge provided by LLMs, enhancing the accuracy of domain-specific information and thus enabling LLMs to generate more precise textual descriptions for molecular samples.
Subsequently, we employ a multi-modal alignment method to coordinate various modalities, including molecular graphs and their corresponding descriptive texts, to guide the pre-training of molecular representations. Extensive experiments demonstrate the effectiveness of the proposed method.


\end{abstract}



\section{Introduction}

Molecular representation learning is a fundamental and preliminary task in the field of drug discovery~\cite{zhang2021mg,wang2022chemical,li2022kpgt,wang2023automated,zhang2024sculpting}. It serves as the basis for following tasks, including molecular property prediction, molecular scaffold optimization, and targeted molecular generation, all of which require robust and expressive molecular representations. Over the years, deep learning techniques have significantly advanced this domain, offering promising avenues for developing accurate and efficient models. 
Molecules, often depicted as graphs, feature nodes representing atoms and edges representing chemical bonds \cite{li2022kpgt}. 
Consequently, graph neural networks (GNNs), as a sophisticated and efficacious framework for graph representation learning~\cite{zhou2020graph,ren2020adversarial,ren2019heterogeneous}, have garnered significant attention and found extensive application in the domain of molecular representation learning~\cite{gasteiger2019directional,jiang2021could}.


However, akin to other deep learning methods, GNN-based approaches are often constrained by the availability of labeled data. Hence, researchers are exploring self-supervised methods for molecular representation learning \cite{rong2020self, liu2021pre, zaidi2022pre, wang2022molecular, fang2023knowledge,ren2021label}.
These self-supervised methods typically follow a paradigm where models are first pre-trained using a large amount of unlabeled molecular samples and then fine-tuned for downstream tasks such as molecule property prediction. The adoption of these pre-trained deep learning models have gained widespread adoption due to their capacity to leverage large-scale unlabeled data for improved performance.

Recent research has witnessed a notable trend towards integrating prior biological domain knowledge into the pre-trained frameworks \cite{yang2021deep, fang2022molecular, fang2023knowledge,zhang2024sculpting}. Incorporating biomedical domain knowledge into the pre-training process enhances the models' ability to learn representations for molecules, resulting in remarkable achievements in molecular property prediction tasks. 
Despite the advancements of these pre-trained frameworks with domain knowledge, a significant bottleneck in this approach is the heavy reliance on biochemical experts to acquire specialized domain knowledge. This process is not only time-consuming but also entails substantial expenses. 
To address this challenge, recent advancements in Natural Language Processing (NLP) have revealed the potential of Large Language Models (LLMs) to efficiently provide knowledge. 
Leveraging this capability, some very recent studies have begun exploring the integration of LLMs into graph domain~\cite{yu2023empower,sun2023large} including molecular tasks~\cite{balaji2023gpt,qian2023can,guo2024moltailor}.

Nevertheless, LLMs for general purpose occasionally exhibit hallucinations and lack precision in generating domain-specific knowledge. For example, an LLM may generate incorrect specialized knowledge or encounter difficulties with mathematical calculations. Thus, the existing methods for introducing LLMs to address molecular tasks may suffer from this drawback. 
In contrast, Domain-Specific Small Models (DSMs) have extensive domain knowledge and can accurately compute domain-related metrics for molecules. Tools such as the RDKit package, a basic type of DSMs, provide specialized calculations. However, due to their limited size and singular functionality, DSMs lack the breadth of knowledge required for comprehensive representation learning.

 Therefore, to leverage the advantages of both LLMs and DSMs, we propose a novel \textbf{Mol}ecular \textbf{Graph} representation learning framework which integrates \textbf{Lar}ge language models and \textbf{Do}main-specific small models, called \textbf{MolGraph-LarDo}. Specifically, we design a two-stage prompt strategy where DSMs are introduced to calibrate the knowledge provided by LLMs, aiming to obtain more accurate domain-specific information and thus enabling LLMs to generate more precise textual descriptions of molecular samples. To further avoid hallucination issues, the designed prompt incorporates both dataset-specific and sample-specific information. 
Subsequently, a multi-modal graph-text alignment method is employed in MolGraph-LarDo to guide the pre-training of molecular graphs. The text modality is informed by textual knowledge from LLMs, while the graph modality originates from the molecular graph structure.
Our experimental results underscore the effectiveness of MolGraph-LarDo in improving the performance of the downstream molecular property prediction while reducing the cost of obtaining specialized domain knowledge.

The contributions of this work are summarized as follows:
\begin{itemize}
\item 
The proposed method leverages the retrieval and generation capabilities of LLMs to overcome the time-consuming and labor-intensive process of biomedical domain literature screening and pre-processing in molecular representation learning.
\item Existing methods that integrate general LLMs into molecular tasks may suffer from hallucinations and lack precision in generating domain-specific knowledge. To address these issues, we introduce a novel framework for molecular graph representation learning which integrates LLMs and DSMs (MolGraph-LarDo).
\item We demonstrate the effectiveness of our proposed method through extensive experiments. Our research focuses on leveraging both LLMs and DSMs to advance molecular representation studies, which could also benefit other LLM-related approaches in broader research areas.
\end{itemize}


\section{Related Work} 
\subsection{Molecular Representation Learning}


Molecular representation learning has witnessed significant advancements in recent years, particularly with methods involving pre-training frameworks.
GROVER \cite{rong2020self} utilizes the message passing networks with a Transformer-style architecture to pretrain models with unlabeled molecular data.
GraphMVP \cite{liu2021pre} employs self-supervised and pre-training strategies to pretrain models using 3D geometric information.
PhysChem~\cite{yang2021deep} is a deep learning framework designed to learn molecular representations using external physical and chemical information.
MG-BERT~\cite{zhang2021mg} is a BERT-based pre-training framework that relies on large amounts of unlabeled data for molecular representation learning.
MGSSL~\cite{zaidi2022pre} is a denoising-based pre-training technique for molecular data.
MolCLR~\cite{wang2022molecular} is a self-supervised framework that leverages large amounts of unlabeled data to pretrain molecules through contrastive learning with GNNs.
KANO~\cite{fang2023knowledge} integrates domain-specific knowledge graphs into the pre-training process of molecular representation learning. 
Although these pre-training methods have achieved good performance in molecular representation learning, they rely on extensive specialized knowledge, including large amounts of unlabeled data and external domain knowledge, which can be costly to obtain.

\subsection{Large Language Models for Molecular Task}
%

In recent years, there has been a surge of interest in leveraging large language models (LLMs) for molecular tasks, as evidenced by some review papers and models in the field.
Researchers provide a comprehensive review~\cite{jin2023large} of LLMs on graphs, detailing the graph scenarios suitable for LLMs and comparing methods within each scenario.
Also, researchers~\cite{zhang2024future} explore potential directions for future research in molecular science using LLMs.
GPT-MolBERTa~\cite{balaji2023gpt} is a BERT-based model that predicts molecular properties using textual descriptions generated by ChatGPT.
LLM4mol~\cite{qian2023can} leverages both in-context classification results and the LLM’s generation of new representations to enhance molecular property prediction.
MolTailor~\cite{guo2024moltailor} optimizes representations by emphasizing task-relevant features, with virtual task descriptions generated by the LLM.
However, existing LLM-related methods often lack precision in generating domain-specific knowledge due to hallucinations. In contrast, the proposed MolGraph-LarDo integrates both LLMs and DSMs to ensure the accuracy of generated domain-specific knowledge.

\begin{figure*}[t]
  \centering
  \includegraphics[width=\textwidth]{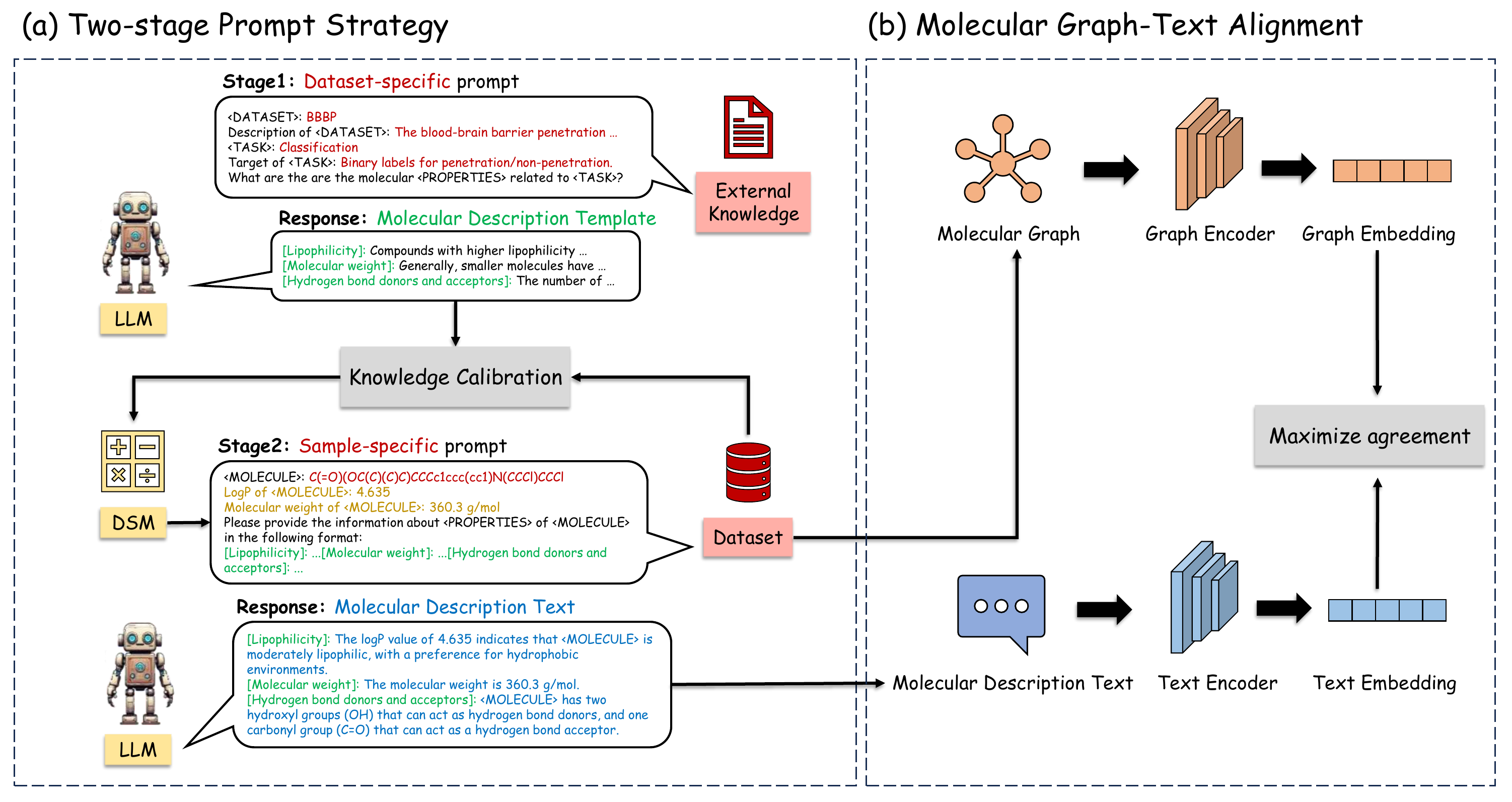}
  \caption{Overview of the proposed framework. (a) Two-stage Prompt Strategy (b) Molecular Graph-Text Alignment.}
  \label{fig:overview}
\end{figure*}

\section{Preliminaries}

Molecular graph representation learning is a preliminary task in molecular property prediction, which can be classified into qualitative molecular property classification and quantitative molecular property regression.

\noindent\textbf{Definition 1.} 
(Molecular Graph):
A molecular graph is a graph $\mathcal{G=(V,E},\mathbf{X})$ representing a molecule, where $\mathcal{V}$ denotes the set of nodes representing atoms, $\mathcal{E}$ represents the set of edges representing chemical bonds, and $\mathbf{X}$ corresponds to the matrix of node features.

\noindent\textbf{Definition 2.} 
(Molecular Property Classification):
Given a set of molecular graphs $\mathbb{G}=\{\mathcal{G}_1, \mathcal{G}_2, \cdots\}$ and a set of classes $\mathcal{C}=\{c_1, c_2, \cdots\}$, the objective of Molecular Property Classification is to learn a mapping function $f:\mathbb{G}\mapsto\mathcal{C}$ that predicts the class $c_i \in \mathcal{C}$ of each molecular graph $\mathcal{G}_i \in \mathbb{G}$.

\noindent\textbf{Definition 3.} 
(Molecular Property Regression): 
Given a set of molecular graphs $\mathbb{G}=\{\mathcal{G}_1, \mathcal{G}_2, \cdots\}$, the objective of Molecular Property Regression is to learn a mapping function $f:\mathbb{G}\mapsto\mathbb{R}$ that predicts the scalar score $s_i \in \mathbb{R}$ of each molecular graph $\mathcal{G}_i \in \mathbb{G}$.

\section{Methodology}

In this section, we present how to utilize both LLMs and DSMs to enhance graph contrastive learning for molecular representation learning. 
In Section~\ref{method:overview}, we first introduce the overview of the proposed MolGraph-LarDo. Section~\ref{method:two-stage} shows the proposed two-stage prompt strategy in detail. The DSM is introduced in this step to implement knowledge calibration.
Section~\ref{method:alignment} elaborates the molecular graph-text alignment, which helps to inject the domain knowledge from LLMs and DSMs into the process of graph contrastive learning.

\subsection{Overview} \label{method:overview}

Figure~\ref{fig:overview} illustrates the overview of the proposed MolGraph-LarDo, which mainly comprises two parts: the two-stage prompt strategy and the molecular Graph-Text Alignment.

Firstly, we design a two-stage prompt strategy to extract domain knowledge with respect to the given molecule. As the name suggested, it consists of two stages.
In Stage 1, we use a dataset-specific prompt, as presented in Section~\ref{method:stage1}, to generate relevant molecular properties to the given dataset. We term these properties as the Molecular Description Template (MD-Template) since they could be included as part of the prompt for generating molecular description. 
We use the DSM to implement knowledge calibration, such as employing RDKit to calculate molecular domain-related metrics for a given molecule. The output of the DSM is termed Calibrated Knowledge.
In Stage 2, we use both the MD-Template and the Calibrated Knowledge to design a sample-specific prompt, as introduced in Section~\ref{method:stage2}. Being fed with the sample-specific prompt, LLM is capable of providing Molecular Description Text (MD-Text), which contains specific information about the dataset-relevant properties of a given molecule. 

Secondly, we introduce a multi-modal alignment method, as presented in Section~\ref{method:alignment}, to enhance contrastive learning by aligning the molecular graph to its corresponding description text. Specifically, MD-Texts generated by LLM are fed into a text encoder to derive text embeddings representing the molecule's description. Similarly, the molecular graphs are inputted into a graph encoder to obtain graph embeddings. Given a molecule in the dataset, its molecular graph embeddings and corresponding description text embeddings are considered as positive pairs in contrastive learning, aiming to optimize their agreement. After the pre-training, the pretrained graph encoder can be used as a model for downstream molecular tasks.

\subsection{The Two-stage Prompt Strategy} \label{method:two-stage}

\begin{figure*}[h]
  \centering
  \includegraphics[width=\textwidth]{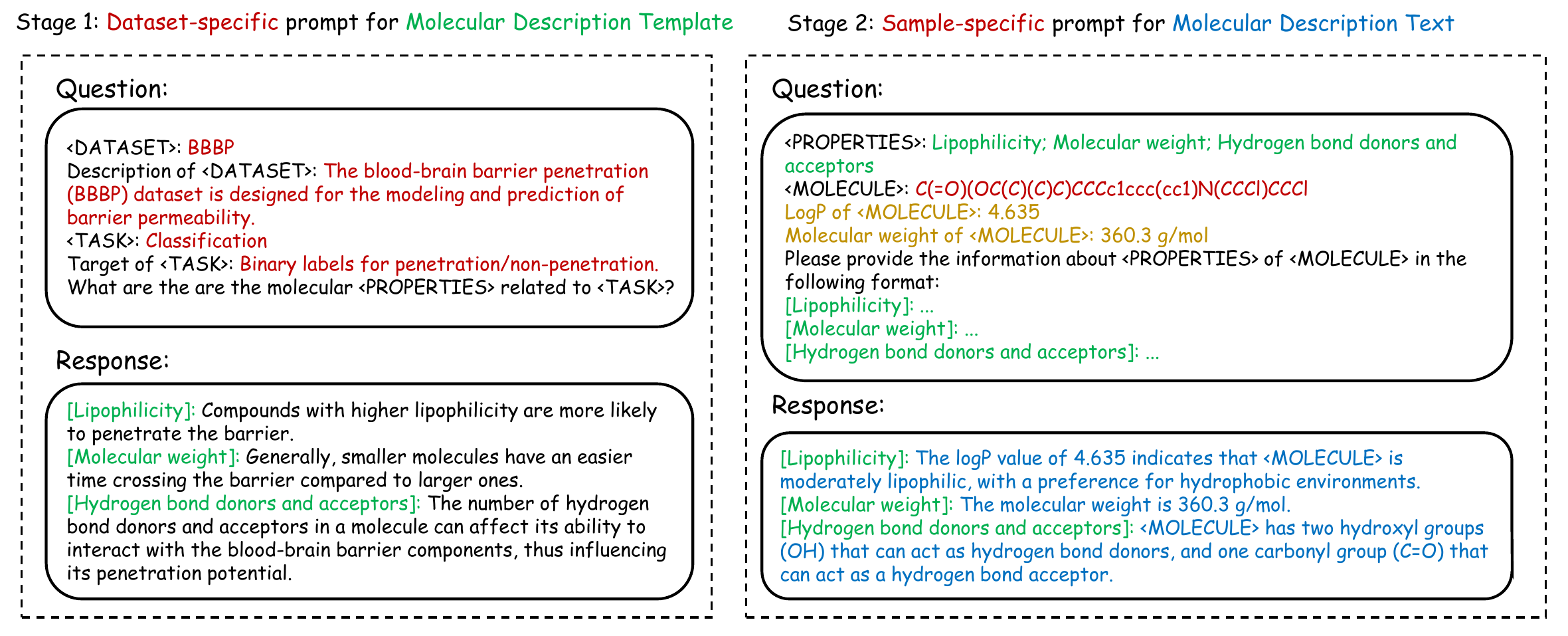}
  \caption{A detailed case of the two-stage prompt strategy on DATASET BBBP. Left: Dataset-specific prompt for generating Molecular Description Template (MD-Template); Right: Sample-specific prompt for generating Molecular Description Text (MD-Text). }
  \label{fig:two-stage}
\end{figure*}

In this part, we aim to employ LLMs and DSMs to provide domain knowledge about a given molecule for the subsequent molecular pre-training. As shown in Figure~\ref{fig:two-stage}, we elaborate the proposed the two-stage prompt strategy with a detailed case as follows.

\subsubsection{\textbf{Stage 1: Dataset-specific Prompt for MD-Template Generation}} \label{method:stage1}
In this stage, we design a dataset-specific prompt to generate several relevant molecular properties, which we term as MD-Template. Naturally, this template is unique and consistent for each dataset. In the following stage, the sample-agnostic template will be used to direct the generation of MD-Text, as shown in Section~\ref{method:stage2}. This process serves to improve the relevance of the generated texts to the specific dataset under consideration.

In the left part of Figure~\ref{fig:two-stage}, we illustrate the dataset-specific prompt for BBBP, a benchmark dataset sourced from MoleculeNet~\cite{wu2018moleculenet}. This involves collecting detailed information about the given dataset, including its name, description, task type, and target variable. We then integrate this information into the prompt and organize it using a predefined format for clarity and consistency. It is noteworthy that the specific details about the dataset are obtained from external sources of information. This process can be viewed as a basic application of Retrieval Augmented Generation (RAG), a technology designed to mitigate the problem of hallucination in LLM~\cite{huang2023survey}. For example, in cases where an LLM may not clearly understand the term "BBBP" based solely on its four-letter format, integrating RAG in the prompt allows us to provide additional external knowledge about the dataset. This approach effectively mitigates hallucination, thereby enhancing the correctness of the generated content.

After inputting the dataset-specific prompt, a LLM would generate several molecular properties along with explanations of their relevance to the dataset. Based on the case illustrated in Figure~\ref{fig:two-stage}, the LLM suggests that "Lipophilicity" is a relevant property concerning the dataset "BBBP," which stands for blood-brain barrier penetration, stating that "Compounds with higher lipophilicity are more likely to penetrate the barrier." Consequently, we collect these properties as MD-Template and include it as part of the prompt in the following stage.

\subsubsection{\textbf{Domain-specific Small Models for Knowledge Calibration}} \label{method:dsm}
LLMs designed for general purposes can occasionally produce inaccurate domain-specific information and struggle with precise calculations. To address this, we introduce Domain-Specific Small Models (DSMs) within our framework to enhance the accuracy of molecular textual information. We refer to this process as knowledge calibration. The necessity of knowledge calibration is demonstrated in the experimental part in Section~\ref{exp:md-text}.

Specifically, we utilize the RDKit package, a basic type of DSMs, to calculate domain-specific metrics for the given molecule, such as "Molecular weight" and "LogP". These metrics are selected from the MD-Template. The output results from the DSM are saved in a specific format, for example, "LogP of \textless MOLECULE \textgreater: 4.635". We refer to these results as calibrated knowledge, which will be incorporated into the prompt for the subsequent stage.

\subsubsection{\textbf{Stage 2: Sample-specific Prompt for MD-Text Generation}} \label{method:stage2}
In this stage, the MD-Template obtained in Section~\ref{method:stage1} and the calibrated knowledge from Section~\ref{method:dsm} are incorporated into the sample-specific prompt. This prompt is then employed to generate MD-Text for each molecule in the dataset. The MD-Text provides molecular textual information and is beneficial for the subsequent graph contrastive learning, as elaborated later in Section~\ref{method:alignment}.

As illustrated in the right part of Figure~\ref{fig:two-stage}, a sample-specific prompt consists of the MD-Template, the calibrated knowledge, and the SMILES notation of a molecule sample. Since the MD-Template contains properties relevant to the given dataset, we anticipate the LLM to generate detailed information regarding these properties for the specified molecule sample.
In this example, we query for the detailed information regarding relevant properties including "Lipophilicity", "Molecular weight" and "Hydrogen bond donors and acceptors" for the molecule "C(=O)(OC(C)(C)C)CCCc1ccc(cc1)N(CCCl)CCCl". In the right part of Figure~\ref{fig:two-stage}, the MD-Text includes the biochemical information about "Lipophilicity", "Molecular weight" and "Hydrogen bond donors and acceptors". The MD-Text of molecules serves as supplementary information for the original dataset, akin to a form of data augmentation. This MD-Text will be preserved for future use, contributing to the graph contrastive learning process in the proposed framework. The integration of LLMs and DSMs provides specialized domain knowledge efficiently, which is invaluable, especially when considering the expert manpower required to accomplish the same task manually.

\subsection{Molecular Graph and MD-Text Alignment} \label{method:alignment}

In this part, we specifically introduce how to utilize the MD-Text obtained from Section~\ref{method:stage2} to empower molecular graph pre-training based on contrastive learning. 
For each sample, it has its own graph structure and MD-Text. Here, the graph structure is sourced from the dataset, while the MD-Text is generated by the LLM and the DSM using the the two-stage prompt strategy.
Considering that each sample now possesses multi-modal information (graph and text), inspired by CLIP~\cite{radford2021learning}, we align the graph representation and text representation of the same sample by maximizing their agreement.
The underlying intuition is that graph representations of molecules with similar textual information about their properties should also be similar, as reflected by MD-text.



Let $\mathcal{G}$ represent the set of all graphs and $\mathcal{T}$ denote the set of all texts. For each sample $i$, its molecular graph $g_i \in \mathcal{G}$ is processed by a graph encoder $f_\text{g}(\cdot):\mathcal{G} \mapsto \mathbb{R}^{d_1}$. Similarly, the MD-Text $t_i \in \mathcal{T}$ associated with sample $i$ is processed by a text encoder $f_\text{t}(\cdot):\mathcal{T} \mapsto \mathbb{R}^{d_2}$. Subsequently, the output embeddings from $f_\text{g}(\cdot)$ and $f_\text{t}(\cdot)$ are projected to the same dimension for joint training. Using the aforementioned notation, the computation of the joint multi-modal embeddings $\mathbf{h}$ can be formulated as follows:
\begin{gather}
\mathbf{h}^\text{g}_i=\mathbf{W}_\text{g}f_\text{g}(g_i) \\
\mathbf{h}^\text{t}_i=\mathbf{W}_\text{t}f_\text{t}(t_i)
  \label{eq:encoder}
\end{gather}
where $\mathbf{W}_\text{g} \in \mathbb{R}^{d \times d_1}$ is the transformation matrix to project the output embedding of the graph encoder into the joint multi-modal embedding space. Similarly, $\mathbf{W}_\text{t} \in \mathbb{R}^{d \times d_2}$ serves the same purpose for the output embedding of the text encoder. 
In practice, the graph encoder is typically a GNN, while the text encoder is usually an NLP-based sequence model. We will describe the specific types of encoders used in this paper later in the Section~\ref{imp}.

\begin{table*}[t]
  \centering
  \caption{Performance comparison between MolGraph-LarDo and baselines. The best performance is \textbf{bolded}.} 
    \begin{tabular}{l|ccc|cccc}
    \toprule
    Task  & \multicolumn{3}{c|}{Classification~$\uparrow$} & \multicolumn{4}{c}{Regression~$\downarrow$} \\
    \midrule
    Datasets & BBBP  & BACE  & SIDER & FreeSolv & ESOL  & Lipo  & QM7 \\
    \midrule
    \# Molecules & 2,039 & 1,513 & 1,427 & 642   & 1,128 & 4,200 & 6,830 \\
    \# Tasks & 1     & 1     & 27    & 1     & 1     & 1     & 1 \\
    \midrule
    GCN   & 0.6438±0.0110 & 0.7775±0.0302 & 0.6016±0.0356 & 2.8823±0.3442 & 1.5573±0.0505 & 0.8427±0.0239 & 89.72±6.72 \\
    GIN   & 0.7002±0.0051 & 0.8231±0.0141 & 0.6050±0.0265 & 2.7930±0.3982 & 1.3705±0.0134 & 0.8090±0.0198 & 85.90±12.96 \\
    \midrule
    Hu et al.   & 0.7082±0.0095 & 0.8124±0.0195 & 0.6048±0.0312 & 2.7623±0.1320 & 1.3213±0.0421 & 0.7622±0.0233  & 88.45±10.36 \\
    GROVER   & 0.7204±0.0096 & 0.8213±0.0271  & 0.6071±0.0124 & 3.1023±0.2372 & 1.4278±0.0512 & 0.8132±0.0363 & 80.23±7.66 \\
    MolCLR-GIN & 0.7154±0.0092 & 0.8164±0.0075 & 0.6009±0.0276 & 2.4489±0.2979 & 1.2954±0.0113 & 0.7486±0.0252 & 86.40±13.72 \\
    \midrule
    \textbf{MolGraph-LarDo} & \textbf{0.7317±0.0152} & \textbf{0.8305±0.0116} & \textbf{0.6095±0.0232} & \textbf{2.3561±0.0789} & \textbf{1.2765±0.0535} & \textbf{0.7482±0.0170} & \textbf{74.91±5.84} \\
    \bottomrule
    \end{tabular}%
  \label{tab:main}%
\end{table*}%

\begin{table*}[t]
  \centering
  \caption{Results of ablation experiments for the two-stage prompt strategy and domain-specific small models.}
    \begin{tabular}{l|ccc|cccc}
    \toprule
    Task  & \multicolumn{3}{c|}{Classification~$\uparrow$} & \multicolumn{4}{c}{Regression~$\downarrow$} \\
    \midrule
    Datasets & BBBP  & BACE  & SIDER & FreeSolv & ESOL  & Lipo  & QM7 \\
    \midrule
    \# Molecules & 2,039 & 1,513 & 1,427 & 642   & 1,128 & 4,200 & 6,830 \\
    \# Tasks & 1     & 1     & 27    & 1     & 1     & 1     & 1 \\
    \midrule
    - w/o TPS \& DSM & 0.6949±0.0190 & 0.8297±0.0187 & 0.5955±0.0259 & 3.3795±1.3439 & 1.3000±0.0451 & 0.7498±0.0129 & 77.39±4.75 \\
    - w/o DSM & 0.7295±0.0031 & 0.8135±0.0482 & 0.5962±0.0349 & 2.6316±0.2328 & 1.3021±0.0076 & 0.7498±0.0059 & 80.32±9.94 \\
    \textbf{MolGraph-LarDo} & \textbf{0.7317±0.0152} & \textbf{0.8305±0.0116} & \textbf{0.6095±0.0232} & \textbf{2.3561±0.0789} & \textbf{1.2765±0.0535} & \textbf{0.7482±0.0170} & \textbf{74.91±5.84} \\
    \bottomrule
    \end{tabular}%
  \label{tab:ablation1}%
\end{table*}%

In line with CLIP, we devise a symmetric loss function to align the graph embedding and text embedding of each sample. Specifically, we compute the batch-wise loss $\mathcal{L}\text{g}$, where the text embedding $\mathbf{h}^\text{t}_i$ of the current sample serves as positive contrast and text embeddings $\mathbf{h}^\text{t}_j$ of other samples in the batch are regarded as negative contrast. Similarly, $\mathcal{L}\text{t}$ is determined in the same manner, except that the positions of the graph and text are interchanged. The calculation of $\mathcal{L}_\text{g}$ and $\mathcal{L}_\text{t}$ could be formulated as follows: 
\begin{gather}
\mathcal{L}_\text{g} = -\frac{1}{N}\sum_{i=1}^{N} \text{log}\frac{\text{exp}(\mathbf{h}_i^\text{g} \cdot \mathbf{h}_i^\text{t} / \tau)}{\text{exp}(\mathbf{h}_i^\text{g} \cdot \mathbf{h}_i^\text{t} / \tau) + \sum_{j \neq i} \text{exp}(\mathbf{h}_i^\text{g} \cdot \mathbf{h}_j^\text{t} / \tau)} \\
\mathcal{L}_\text{t} = -\frac{1}{N}\sum_{i=1}^{N} \text{log}\frac{\text{exp}(\mathbf{h}_i^\text{t} \cdot \mathbf{h}_i^\text{g} / \tau)}{\text{exp}(\mathbf{h}_i^\text{t} \cdot \mathbf{h}_i^\text{g} / \tau) + \sum_{j \neq i} \text{exp}(\mathbf{h}_i^\text{t} \cdot \mathbf{h}_j^\text{g} / \tau)}
  \label{eq:loss_gt}
\end{gather}
where $\mathbf{h}^\text{g}_i$ and $\mathbf{h}^\text{t}_i$ represent the graph and text embeddings of the $i$-th sample respectively; $N$ is the batch size; $\tau$ is the temperature parameter of InfoNCE~ \cite{oord2018representation}. Then the final loss is calculated by averaging $\mathcal{L}_\text{g}$ and $\mathcal{L}_\text{t}$ as:
\begin{equation}
  \mathcal{L}=\frac{1}{2} (\mathcal{L}_\text{g}+\mathcal{L}_\text{t})
\end{equation}

For optimization, we update the weights of graph encoder by minimizing $\mathcal{L}$. After pre-training, the pretrained graph encoder can be fine-tuned for downstream molecular tasks.

\section{Experimental Settings}
\subsection{Datasets}

To evaluate the proposed MolGraph-LarDo, we conduct experiments over seven datasets sourced from MoleculeNet~\cite{wu2018moleculenet}. These molecular datasets can be classified into two types based on task: classification and regression. 

\subsubsection{\textbf{Classification Datasets}}

 \textbf{BBBP} dataset tracks whether a molecule can penetrate the blood-brain barrier and includes binary labels for over 2,000 molecules.
\textbf{BACE} dataset contains qualitative binary labels of 1,513 molecules that inhibit human $\beta$-secretase 1 (BACE-1)
 \textbf{SIDER} dataset includes data on 1,427 approved drug molecules, with side effects categorized into 27 system organ classes according to MedDRA classifications.

\subsubsection{\textbf{Regression Datasets}}

\textbf{FreeSolv} dataset includes hydration free energies of small molecules in water, derived from both experimental measurements and alchemical free energy calculations. 
\textbf{ESOL} dataset records the water solubility of 1,128 molecules.
\textbf{Lipo} dataset measures the lipophilicity of around 4,200 drug molecules, which impacts both membrane permeability and solubility. 
\textbf{QM7} is a dataset derived from the GDB-13 database, containing 6,830 molecules. It is utilized for studying molecular structures and their atomization energies.


\subsection{Baselines}

To demonstrate the superiority of the proposed MolGraph-LarDo, we compare it with five baseline methods, which can be classified into two types as follows.

\subsubsection{\textbf{Supervised Methods}}
\textbf{GCN}~\cite{kipf2016semi} and \textbf{GIN}~\cite{xu2018powerful} are two classic and representative graph neural network models.

\subsubsection{\textbf{Pre-training Methods}}
\textbf{Hu et al.}~\cite{hu2020strategies} introduced a pre-training strategy for GNNs that enhances learning at both individual node and entire graph levels.
\textbf{GROVER}~\cite{rong2020self} integrates the message passing networks with a Transformer-style architecture to pretrain models from unlabelled molecular data.
\textbf{MolCLR}~\cite{wang2022molecular} is a self-supervised framework that leverages large amounts of unlabelled data to pretrain molecules via graph contrastive learning.

\subsection{Implementation Details}
\label{imp}

For the implementation of the proposed MolGraph-LarDo, we use GIN as the graph encoder and Sentence-BERT~\cite{reimers2019sentence} as the text encoder. The GIN consists of five layers with 256 hidden dimensions. The weights of the text encoder are frozen, except for the last output linear layer, which has 256 hidden dimensions. The LLM version used in our experiment is Mistral-7B-Instruct-v0.2, and the DSM version is RDKit-2023.3.2.
Following~\cite{wang2022molecular}, we convert SMILES strings into graphs, where node features are represented by atomic number and chirality, and edge features are represented by bond type and bond direction.

All datasets are split into training, validation, and test sets using a scaffold split method with a ratio of 8:1:1.
In the pre-training stage, we use Adam as the optimizer with a learning rate set to 0.005. The training consists of 100 epochs, including 10 warm-up epochs, and uses a batch size of 32. The model’s pretrained weights are saved when the lowest validation loss is encountered.
In the fine-tuning stage, we search the learning rate within [0.0001, 0.0005] for the fine-tuned graph encoders with a maximum of 100 epochs. The batch size is set to 32 for all datasets except FreeSolv, for which a batch size of 8 is used. 

For evaluation, the ROC-AUC metric is used for classification datasets, whereas RMSE is employed for all regression datasets except for QM7, for which MAE is used. 
We repeat the experiments three times and report the mean and standard deviation. For multi-task datasets, the results are the average of each task's outcomes.

\section{Experiments and Discussion}

In this section, to verify the effectiveness of the proposed MolGraph-LarDo, we conduct experiments to investigate following research questions:
\begin{itemize}
\item\textbf{RQ1:} How does MolGraph-LarDo perform on molecular property prediction as compared to other baseline methods? (Section~\ref{exp:property-prediction})
\item\textbf{RQ2:} Are the proposed two-stage prompt strategy and domain-specific small models effective in MolGraph-LarDo? (Section~\ref{exp:two-stage})
\item\textbf{RQ3:} What is the quality of the MD-Text generated by MolGraph-LarDo? (Section~\ref{exp:md-text})
\item\textbf{RQ4:} Does the graph-text alignment mechanism of MolGraph-LarDo work as expected? (Section~\ref{exp:alignment})

\end{itemize}
\begin{figure*}[t]
  \centering
  \includegraphics[width=0.95\textwidth]{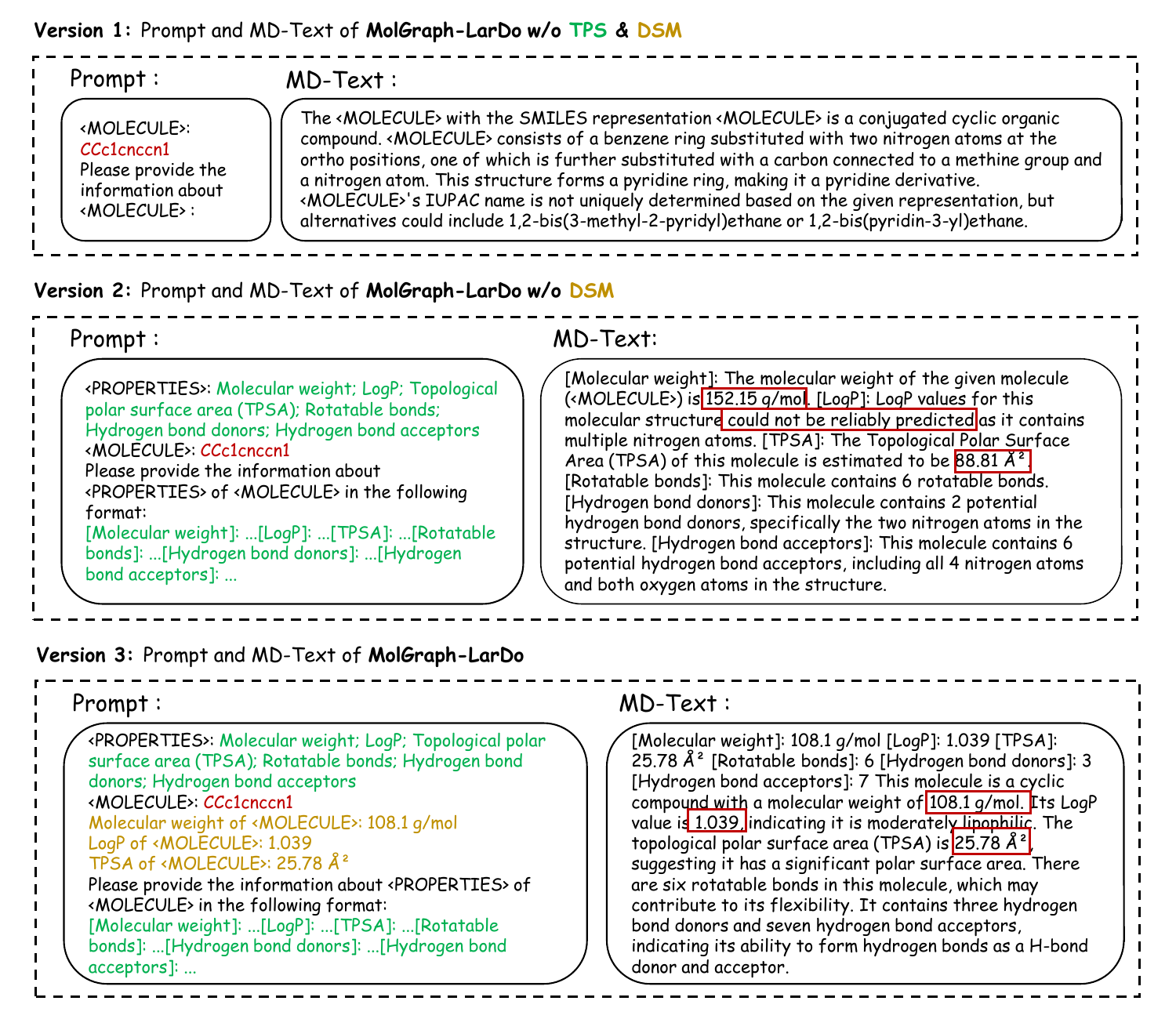}
  \caption{Case study of prompt and MD-Text on FreeSolv dataset for three versions of MolGraph-LarDo.}
  \label{fig:md-text}
\end{figure*}

\subsection{Molecular Property Prediction~(RQ1)}
\label{exp:property-prediction}

In this section, we compare the performance of our proposed method with several baseline methods. As shown in TABLE~\ref{tab:main}, the main observations are as follows: 
(1) The pre-training methods generally outperform the supervised methods in most cases. This advantage likely arises from the pre-training process, which incorporates external knowledge from large amount of unlabeled datasets. 
(2) On the seven datasets, the proposed MolGraph-LarDo consistently outperforms all baseline methods, including those requiring large amounts of domain-specific data for pre-training. Thus, MolGraph-LarDo achieves strong performance without the need for time-consuming and expensive domain knowledge acquisition. This success can be attributed not only to the high efficiency of LLMs in generating data but also to the accuracy of domain knowledge provided by DSMs.

\subsection{Ablation Study: Effect of the Two-stage Prompt Strategy and Domain-specific Small Models~(RQ2)}
\label{exp:two-stage}

To investigate the impact of the designed two-stage prompt strategy and domain-specific small Models, we compare the three versions of MolGraph-LarDo.
We report the results in TABLE~\ref{tab:ablation1}. For clarification, "w/o" stands for "without," while "TPS" and "DSM" are abbreviations for Two-stage Prompt Strategy and Domain-specific Small Models, respectively.
TABLE~\ref{tab:ablation1} shows that MolGraph-LarDo outperforms all other versions, which justifies the rationality of the design of the two-stage prompt strategy and the utilization of Domain-specific Small Models in the framework.


\subsection{Case Study of the MD-Text~(RQ3)}
\label{exp:md-text}

\begin{figure}[t]
  \centering
  \includegraphics[width=\linewidth]{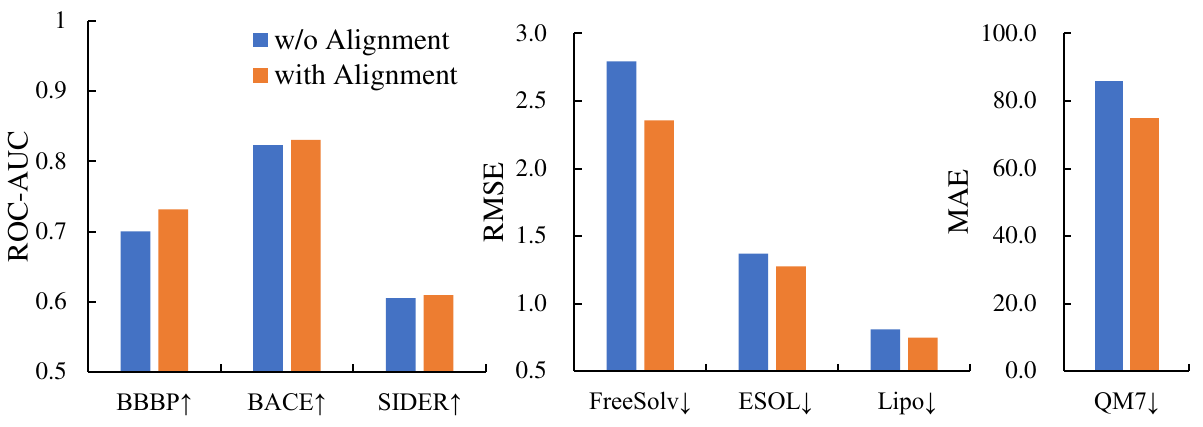}
  \caption{Results of ablation experiments for the graph-text alignment.}
  \label{fig:ablation}
\end{figure}

To more clearly demonstrate the differences between the three versions of MolGraph-LarDo discussed in Section~\ref{exp:two-stage}, we present the case study of both the input prompt and the output MD-Text for each version in Figure~\ref{fig:md-text}.
As the figure shows, Version 1 only generates a general description of the given molecule. Without TPS, it cannot ensure the relevance of the knowledge to specific tasks, and without DSM, it cannot ensure the accuracy of the knowledge.
In contrast, Version 2 introduces TPS, making the generated MD-Text relevant to the given task. However, the accuracy of some domain-related metrics remains insufficient. For instance, several metrics highlighted in the red box yield incorrect answers. This is due to the LLM's poor performance in domain-specific scientific calculations and the accompanying issue of hallucinations.
For Version 3, i.e., MolGraph-LarDo, both TPS and DSM are utilized. They ensure the relevance of the generated MD-Text and also correct the erroneous domain-related metrics from Version 2. For example, it calibrate the molecular weight from the incorrect 152.15 g/mol to the accurate 108.1 g/mol.
This observation is in accord with our design motivation that the DSM are introduced to calibrate the knowledge provided by the LLM, enabling the LLM to produce more accurate textual descriptions of molecular samples.

\subsection{Ablation Study: Effect of the Graph-Text Alignment~(RQ4)}
\label{exp:alignment}

To validate the effectiveness of the proposed graph-text alignment, we conduct experiments fro ablation study. As shown in Figure~\ref{fig:ablation}, MolGraph-LarDo consistently outperforms the version without the graph-text alignment mechanism. 
This observation supports the design idea that graph representations of molecules with similar textual descriptions of properties should also be similar, thereby enhancing performance in downstream molecular tasks.



\section{Conclusion}

LLMs excel at understanding and efficiently providing general knowledge.
Existing methods using general LLMs for molecular tasks may suffer from hallucinations and lack precision in domain-specific knowledge.
To address the issue, we propose a novel framework for Molecular Graph representation learning that integrates LLMs with DSMs (MolGraph-LarDo). Specifically, we develop a two-stage prompt strategy where the DSM calibrates domain-specific knowledge for the LLM, enabling it to produce more accurate textual descriptions of molecular samples. MolGraph-LarDo then employs a multi-modal graph-text alignment method for pre-training molecular graphs. Our results demonstrate that MolGraph-LarDo effectively enhances downstream molecular property prediction while reducing the cost of acquiring domain-specific knowledge used for pre-training. The source code for reproducibility is available at \url{https://github.com/zhangtia16/MolGraph-LarDo}

\bibliographystyle{IEEEtran}
\bibliography{references.bib}











\end{document}